\documentclass[runningheads]{llncs}
\usepackage{graphicx}
\usepackage{amsmath}
\usepackage{hyperref}
\hypersetup{hidelinks,
	colorlinks=true,
	allcolors=black,
	pdfstartview=Fit,
	breaklinks=true}
\usepackage{array}
\newcolumntype{M}[1]{>{\centering\arraybackslash}m{#1}}
\begin{document}
\title{Dewarping Document Image By Displacement Flow Estimation with Fully Convolutional Network}
\titlerunning{Dewarping Document Image By Displacement Flow Estimation}
%
\author{Guo-Wang Xie\inst{1,2} \and
Fei Yin\inst{2} \and
Xu-Yao Zhang\inst{1,2} \and
Cheng-Lin Liu\inst{1,2,3}}
\authorrunning{Xie et al.}
%
\institute{
School of Artificial Intelligence, University of Chinese Academy of Sciences, Beijing 100049, P.R. China\and
National Laboratory of Pattern Recognition, Institute of Automation of Chinese Academy of Sciences, 95 Zhongguancun East Road, Beijing 100190, P.R. China \and
CAS Center for Excellence of Brain Science and Intelligence Technology, Beijing, P.R. China\\
\email{xieguowang2018@ia.ac.cn}\\
\email{\{fyin, xyz, liucl\}@nlpr.ia.ac.cn}}
\maketitle              
\begin{abstract}
As camera-based documents are increasingly used, the rectification of distorted document images becomes a need to improve the recognition performance. In this paper, we propose a novel framework for both rectifying distorted document image and removing background finely, by estimating pixel-wise displacements using a fully convolutional network (FCN). The document image is rectified by transformation according to the displacements of pixels. The FCN is trained by regressing displacements of synthesized distorted documents, and to control the smoothness of displacements, we propose a Local Smooth Constraint (LSC) in regularization. Our approach is easy to implement and consumes moderate computing resource. Experiments proved that our approach can dewarp document images effectively under various geometric distortions, and has achieved the state-of-the-art performance in terms of local details and overall effect. Our code and trained models are available at \href{https://github.com/gwxie/Dewarping-Document-Image-By-Displacement-Flow-Estimation}{https://github.com/gwxie/Dewarping-Document-Image-By-Displacement-Flow-Estimation}.

\keywords{Dewarping Document Image \and Pixel-Wise Displacement \and Fully Convolutional Network \and Local Smooth Constraint.}
\end{abstract}
\begin{figure}
\includegraphics[width=\textwidth]{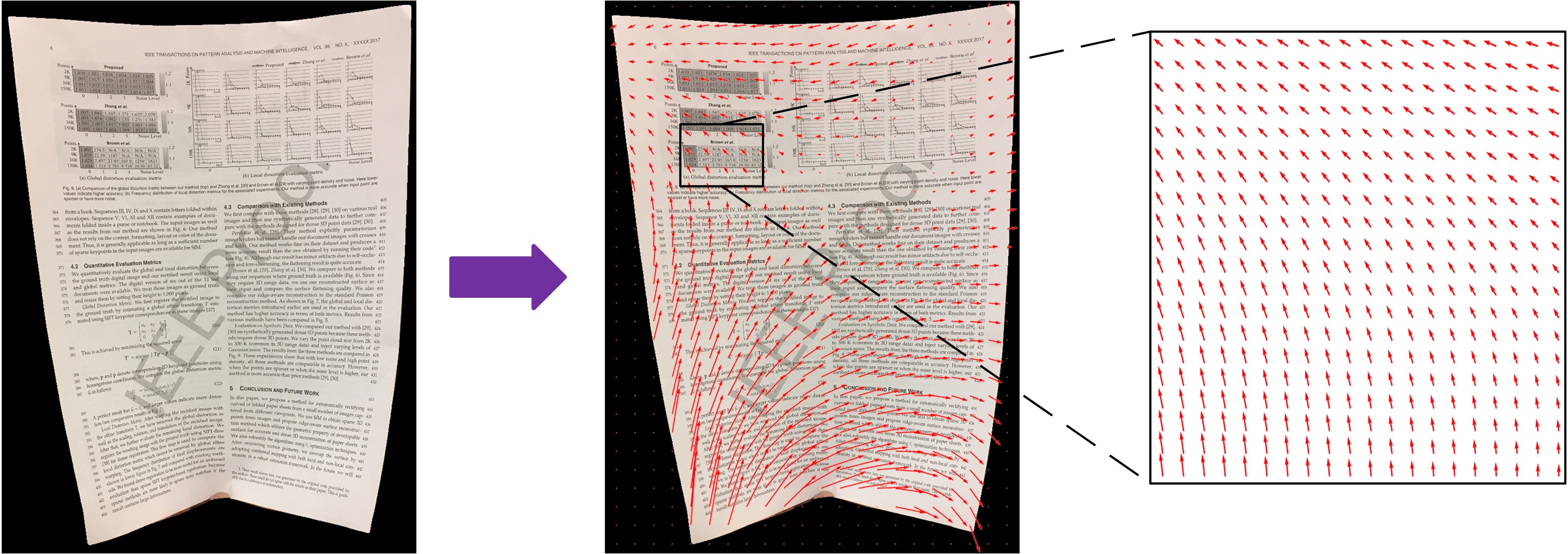}
\caption{Our approach regards the document image as a field of displacement flow, which represents the displacements of pixels for transforming one image into another for rectification.} \label{fig6}
\end{figure}

\section{Introduction}
With the popularity of mobile devices in recent years, camera-captured document images are becoming more and more common. Unlike document images captured by flat scanners, camera-based document images are more likely to deform due to multiple factors such as uneven illumination, perspective change, paper distortion, folding and wrinkling. This makes the processing and recognition of document image more difficult. To reduce the effect of distortion in processing of document images, dewarping approaches have been proposed to estimate the distortion and rectify the document images.

Traditional approaches estimate the 3D shape of document images using auxiliary hardware~\cite{ref_1,ref_2} or the geometric properties and visual cues of the document images~\cite{ref_3,ref_4,ref_5}. Some approaches~\cite{ref_6,ref_7} restrict the page surface to be warped as a cylinder for simplifying the difficulty of 3D reconstruction. Then using raw images and the document shape computer flattened image to correct the distortions. These methods require specific hardware, external conditions or strong assumptions which restrict their generality. For improving the generality of dewarping model, the methods in~\cite{ref_8} and~\cite{ref_9} use deep neural networks to regress the dewarping function from deformed document image by using 2D and 3D supervised information of the warping respectively. Li et al.~\cite{ref_10} considered that it was not possible to accurately and efficiently process the entire image, and proposed patch-based learning approach and stitch the patch results into the rectified document by processing in the gradient domain. Although these methods have obtained promising performance in rectification, further research is needed to deal with situations of more difficult distortions and background.

In this paper, we propose a novel framework to address the difficulties in both rectifying distorted document image and removing background finely. We view the document image is a field of displacement flow, such that by estimating pixel-wise displacements, the image can be transformed to another image accordingly. For rectifying distorted documents, the displacement flow is estimated using a fully convolutional network (FCN), which is trained by regressing the ground-truth displacements of synthesized document images. The FCN has two output branches, for regressing pixel displacements and classifying foreground/background. We design appropriate loss functions for training the network, and to control the smoothness of displacements, we propose Local Smooth Constraint (LSC) for regularization in training. Fig.~\ref{fig6} shows the effect of displacement flow and image transformation. 

Compared with previous methods based on DNNs, our approach is easy to implement. It can process a whole document image efficiently in moderate computation complexity. The design of network output layers renders good effect in both rectifying distortion and removing background. The LSC in regularization makes the rectified image has smooth shape and preserves local details well. Experiments show that our approach can rectify document images and various contents and distortions, and yields state of the art performance on real-world dataset.

\section{Related Works}
A lot of techniques for rectifying distorted document have been proposed in the literature. We partitioned them into two groups according to whether deep learning is adopted or not.

{\bfseries Non-deep-learning-based rectification.} Prior to the prevalence of deep learning, most approaches rectified document image by estimating the 3D shape of the document images. For reconstructing the 3D shape of document image, many approaches used auxiliary hardware or the geometric properties of the document images to compute an approximate 3D structure. Zhang et al.~\cite{ref_1} utilized a more advanced laser range scanner to reconstruct the 3D shape of the warped document. Meng at al.~\cite{ref_2} recovered the document curl by using two structured laser beams. Tsoi et al.~\cite{ref_11} used multi-view document images and composed together to rectify document image. Liang et al.~\cite{ref_6} and Fuet al. ~\cite{ref_7} restricted the page surface to be warped as a cylinder to simplify the difficulty of 3D reconstruction. Moreover, some techniques utilized geometric properties and visual cues of the document images to reconstruct the document surface, such as illumination/shading~\cite{ref_3,ref_4}, text lines~\cite{ref_5,ref_12}, document boundaries~\cite{ref_13,ref_14} etc. Although these method can  handle simple skew, binder curl, and fold distortion, it is difficult for complicated geometric distortion(i.e., document suffer from fold, curve, crumple and combinations of these etc.) and changeable external conditions (i.e., camera positions, illumination and laying on a complex background etc.)

{\bfseries Deep-learning-based rectification.} The emergence of deep learning inspires people to investigate the deep architectures for document image rectification. Das et al.~\cite{ref_15} used a CNN to detect creases of document and segmented document into multiple blocks for rectification.  Xing et al.~\cite{ref_16} applied CNN to estimate document deformation and camera attitude for rectification. Ramanna et al.~\cite{ref_17} removed curl and geometric distortion of document by utilizing a pix2pixhd network (Conditional Generative Adversarial Networks). However, these method were only useful for simple deformation and monotone background. Recently, Ma et al.~\cite{ref_8} proposed a stacked U-Net which was trained end-to-end to predict the forward mapping for the warping. Because of the generated dataset is quite different from the real-world image, ~\cite{ref_8} trained on its dataset has worse generalization when tested on real-world images. Das and Ma et al.~\cite{ref_9} think dewarping model was not always perform well when trained by the synthetic training dataset only used 2D deformation, so they created a Doc3D dataset which has multiple types of pixel-wise document image ground truth by using both real-world document and rendering software. Meanwhile, ~\cite{ref_9} proposed a dewarping network and refinement network to correct geometric and shading of document images. Li et al.~\cite{ref_10} generated training dataset in the 3D space and use rendering engine to get the finer, realistic details of distorted document image. They proposed patch-based learning approach and stitch the patch results into the rectified document by processing in the gradient domain, and a illumination correction network used to remove the shading. Compared to prior approaches, ~\cite{ref_9,ref_10} cared more about the difference between the generated training dataset and the real-world testing dataset, and focused on generating more realistic training dataset to improve generalization in real-world images. Although these results are amazing, the learning and expression capability of deep neural network was not fully explored.

\section{Proposed Approach}
Our approach uses a FCN with two output branches for predicting pixel displacements and foreground/background classification. In dewarping, the foreground pixels are mapped to the rectified image by interpolation according to the predicted displacements.

\subsection{Dewarping Process}
\begin{figure}
\includegraphics[width=\textwidth]{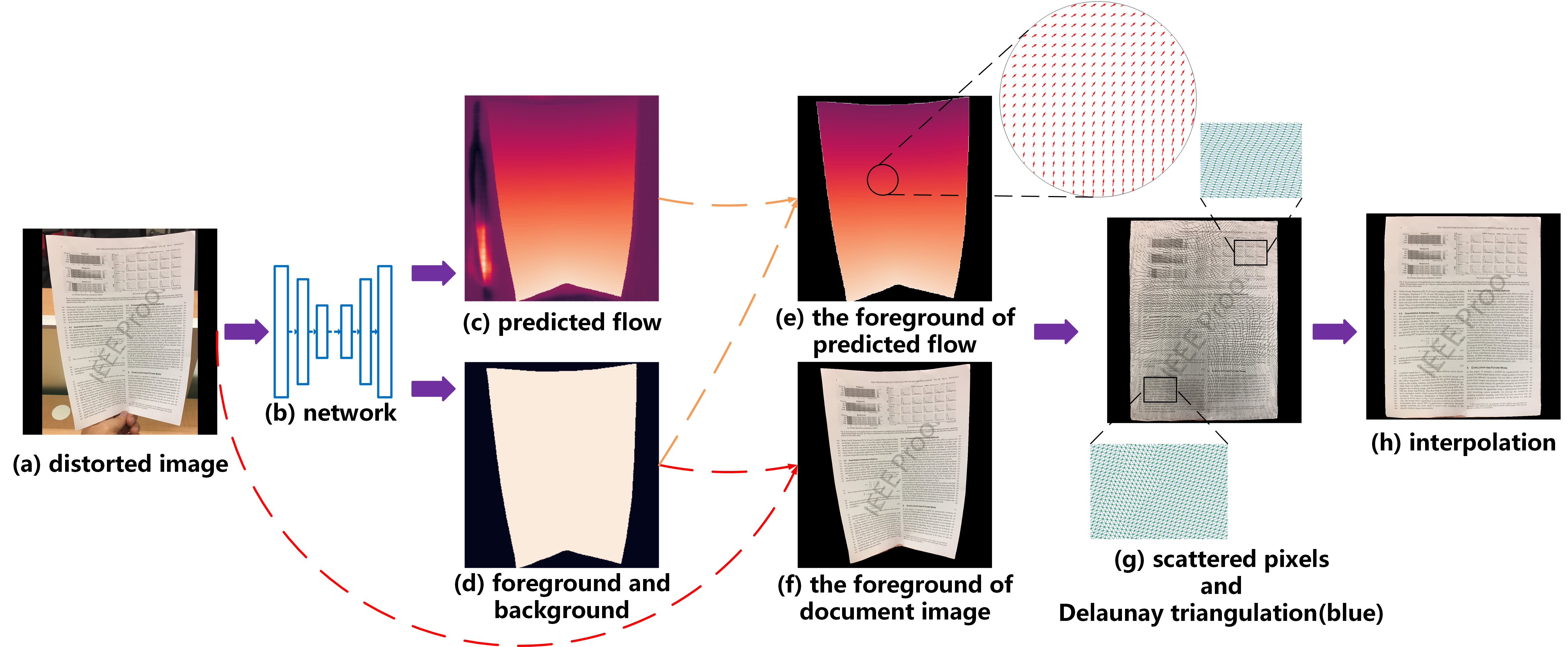}
\caption{Illustration of the process of dewarping document image. An input distorted document is first fed into network to predict the pixel-wise displacements and foreground/background classification. When performing rectification, Delaunay triangulation is applied for interpolation in all scattered pixels.} \label{fig3}
\end{figure}
Fig.~\ref{fig3} illustrates the process of dewarping document image in our work. We predict the displacement and the categories (foreground or background) at pixel-level by applying two tasks in FCN, and then remove the background of the input image, and mapped the foreground pixels to rectified image by interpolation according to the predicted displacements. The cracks maybe emerge in rectified image when using a forward mapping interpolation. Therefore, we construct Delaunay triangulations in all scattered pixels and then using interpolation~\cite{other_1}.

\begin{figure}
\includegraphics[width=\textwidth]{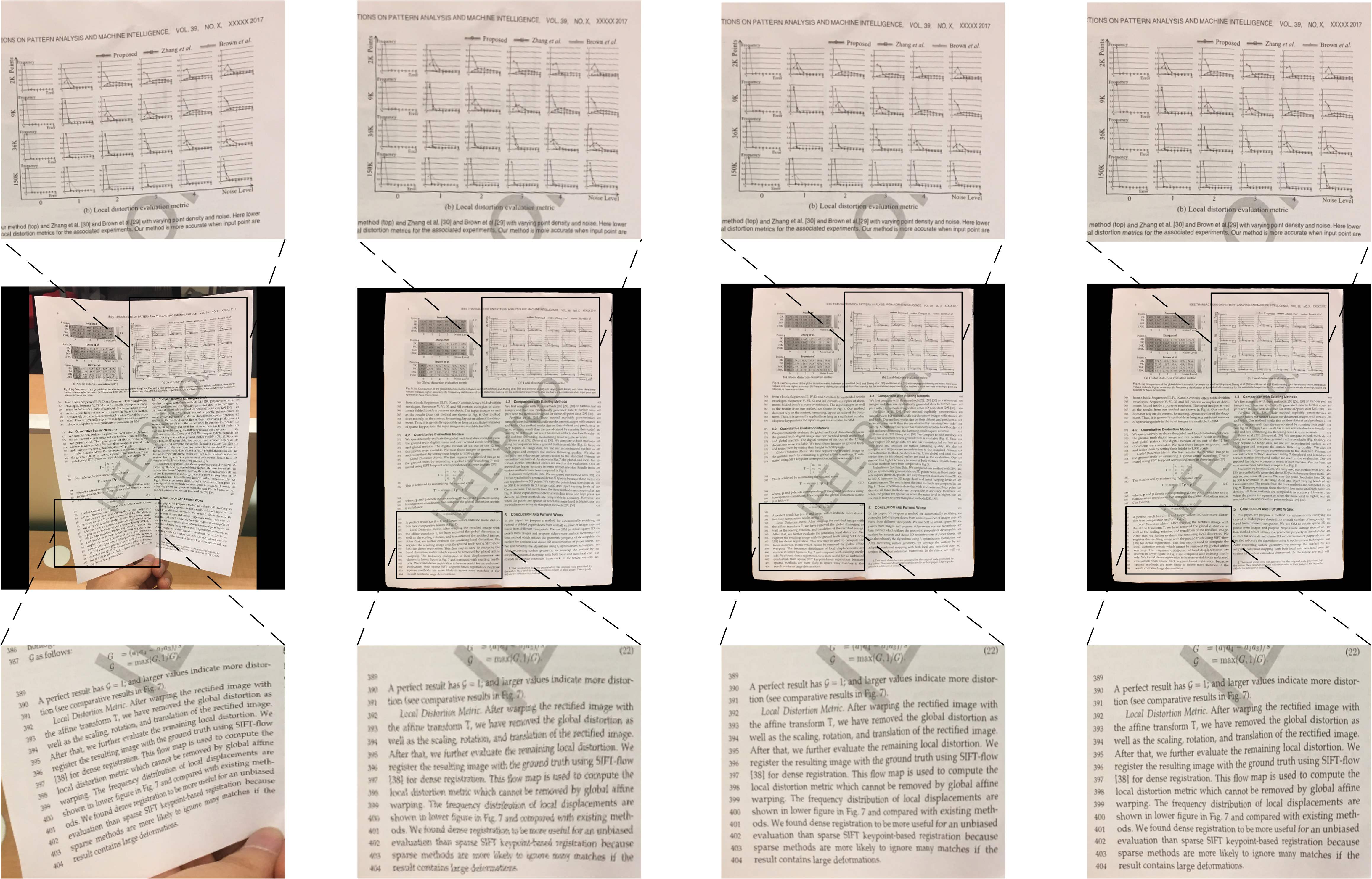}
\caption{Results of the different resolution by zooming the flow of displacement. Column 1 : Original distorted image, Column 2 : Initial (x1) rectified image, Column 3 : x1.5 rectified image, Column 4 : x2 rectified image} \label{fig4}
\end{figure}

For facilitating implementation, we resize the input image into 1024x960 (zooming in or out along the longest side and keeping the aspect ratio, then filling zero for padding. ) in our work. Although smaller input image requires less computing, some information may be lost or unreadable when the distorted document image has small text, picture etc. To trade-off between computational complexity and rectification effect, the document pixels are mapped to rectified image of the same size as the original image, and all the pixels in rectified image are filled by interpolation. We adjust the mapping size as follows:
\begin{equation}
I = F (\lambda \cdot \Re ; I_{HD}) ,
\end{equation}
where $\lambda$ is the scaling factor of zooming in or out, $\Re$ is the map of displacement prediction, $I_{HD}$ is the high-resolution distorted image which has same size as $\lambda \cdot \Re$, $F$ is the linear interpolation and $I$ is the rectified image with higher resolution. As shown in Fig.~\ref{fig4}, we can implement this method when computing and storage resources are limited.

\subsection{Network Architecture}

\begin{figure}
\includegraphics[width=\textwidth]{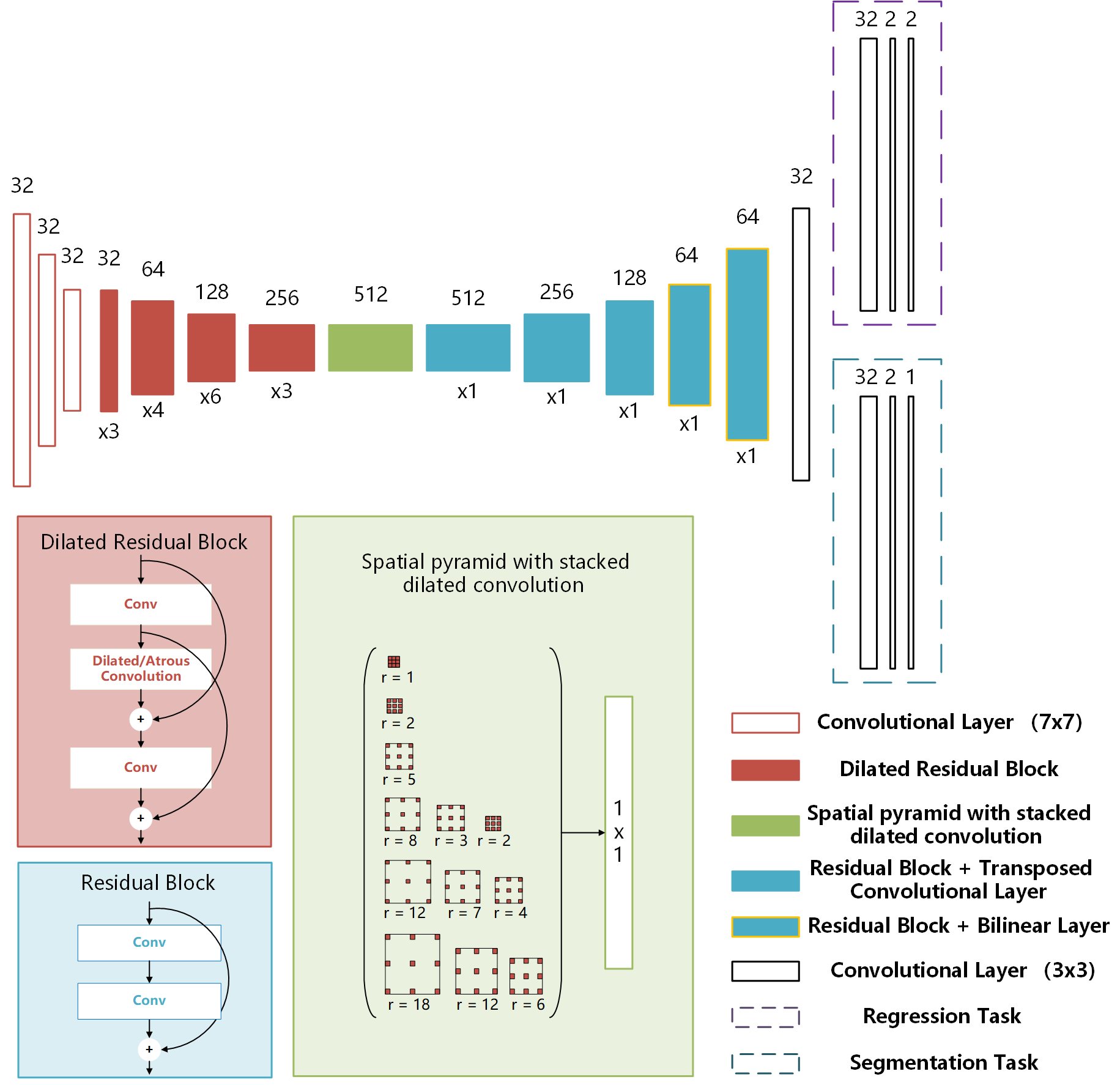}
\caption{Illustration of the FCN architecture. The network has two output branches, for  pixel displacements prediction and foreground/background classification, respectively.} \label{fig1}
\end{figure}

In this section, we introduce the architecture of neural network as shown in Fig.~\ref{fig1}. For improving the generalization in real-world images, instead of focusing on the vulnerable visual cues, such as illumination/shading, clear text etc, our network architecture infer the displacement of entire document from the image texture layout. Compared with ~\cite{ref_8,ref_9}, our method can simplifies the difficulty of rectification because our model need not to predict the global position of each pixel in flatten image. Different from~\cite{ref_10}, our approach can take into account both local and global distortion. 

We adopt an auto-encoder structure and add group normalization (separating the channels into 32 groups) and ReLU after each convolution. To trade-off between computational complexity and rectification effect, the encoder extract local feature by using three convolutional layers with three strides of 1, 2, 2 and 7x7 kernels.  Inspired by the architecture from~\cite{other_3,other_4}, We design a dilated residual block which fuse local and dilated semantic by utilizing general convolution, dilated convolution with rate=3 and residual connection. In this way, we can extract denser and larger receptive field distortion feature. After that, we use one spatial pyramid with stacked dilated convolution to encode global high-level semantic information by parallel and cascaded manners. Distortion feature extractor reduces the spatial resolution and obtain the global feature maps, then we gradually recover the displacement of the entire image (the raw resolution) from the spatial feature by using residual block~\cite{other_2} with transposed convolutional layer or bilinear layer. 

We use multi-task manner to rectify the document image and  separate the foreground and background. The regression task applies group normalization and PReLU after each convolution except for the last layer, and the segmentation task applies group normalization and ReLU after each convolution except for the last layer which adds a sigmoid layer.

\subsection{Loss Functions}
We train the deep neural network by defining four loss function as a guide to regress the compact and smooth displacement and separate the foreground and background.

The segmentation loss we use in this work is the standard cross entropy loss, which is defined as:
\begin{equation}
L_B= -\frac{1}{N} \sum_{i}^{N}\left [ y_i \cdot log\left (  \hat{p_i}\right )+\left (  1-y_i\right ) \cdot log\left (  1-\hat{p_i}\right ) \right ] ,
\end{equation}
where $N$ is the number of elements in flow, $y_i$ and $\hat{p_i}$ respectively denote the ground-truth and predicted classification.

We optimize the network by minimizing the L1 element-wise loss which measures the distance of pixel-displacement of the foreground between the predicted flow and the ground-truth flow. We formulate $L_D$ function as follows: 
\begin{equation}
L_D=\frac{1}{N_f} \sum_{i}^{N_f} \|\Delta D_i - \Delta \hat{D_i}\|_1 ,
\end{equation}
where $N_f$ is the elements of foreground which is specified by ground-truth. $D_i$ and $\hat{D_i}$ denote the pixel-displacement in ground-truth and output value of regression network, respectively.

\begin{figure}
\includegraphics[width=\textwidth]{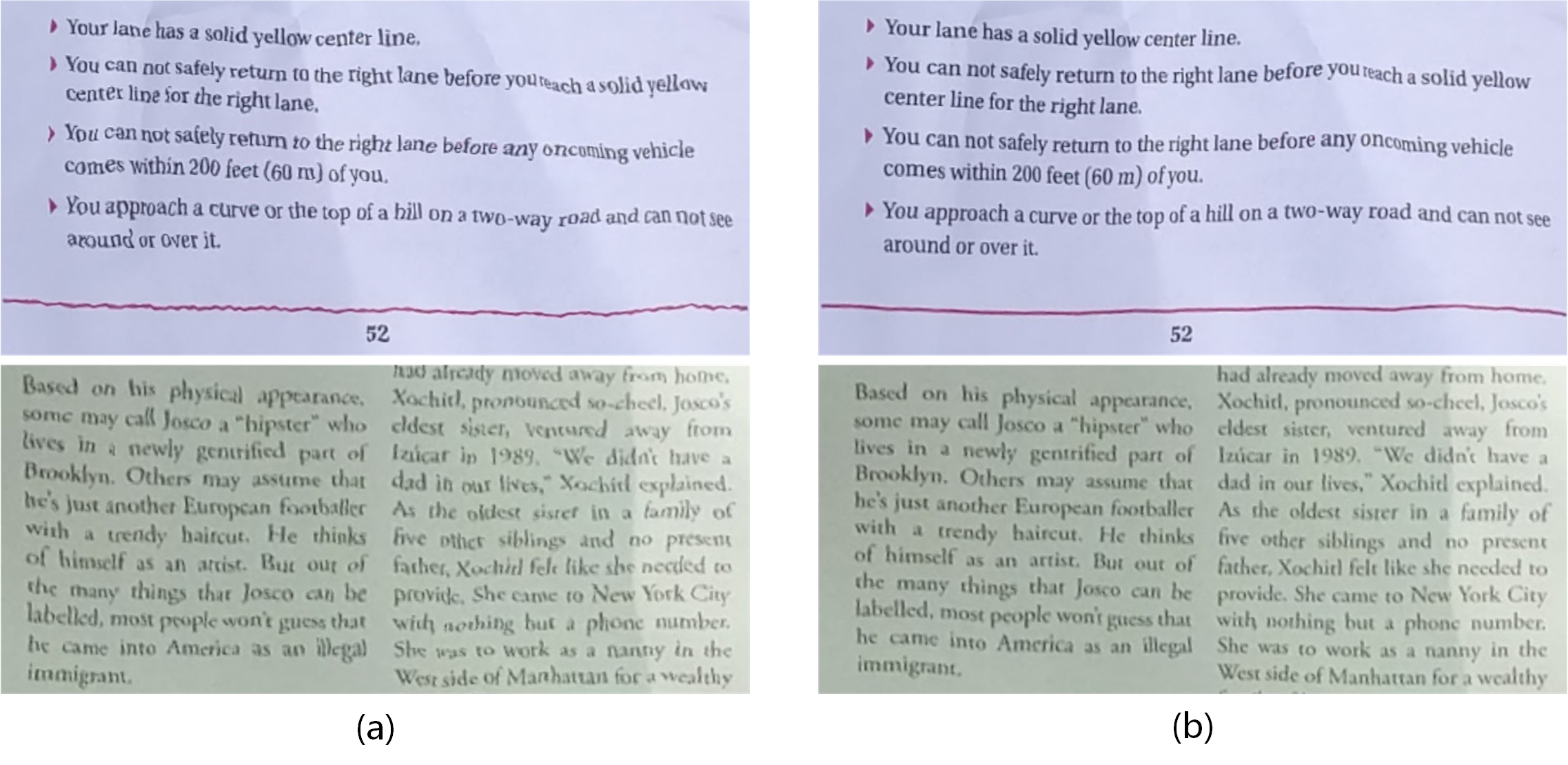}
\caption{Results of using Local Smooth Constraint (LSC) after 8 epoch of training. (a) Without using LSC. (b) Using LSC.} \label{fig5}
\end{figure}

Although the network can be trained by measuring the pixel-wise error between the generated flow and the ground-truth, it's difficult to make model obey the continuum assumption between pixels as shown in  Fig.~\ref{fig5}. To keep the vary continuously from one point to another in a local, we propose a Local Smooth Constraint (LSC). In a local region, the LSC expects the predicted displacement trend to be as close to the ground-truth flow as possible. The displacement trend represents the relative relationship between a local region and its central point, which can be defined as :
\begin{equation}
\delta = \sum_{j=1}^{k}(\Delta D_j - \Delta {D_{center}}) ,
\end{equation}
where $k$ is the number of elements in a local region. In our work, we define the local region as a 3x3 rectangle, and $\Delta {D_{center}}$ represents the center of rectangle. To speed up the calculation, we apply a 2D convolution with a strides of 1 and 3x3 kernel. We formulate LSC as follows:
\begin{equation}
\begin{aligned}
L_{LSC} &= \frac{1}{N_f}\sum_{i}^{N_f} \| \delta_i - \hat{\delta_i}\|_1 \\
&= \frac{1}{N_f}\sum_{i}^{N_f} \| \sum_{j=1}^{k}(\Delta D_j - \Delta {D_{i}}) - \sum_{j=1}^{k}(\Delta \hat{D_j} - \Delta {\hat{D_{i}}})\|_1 \\
&= \frac{1}{N_f}\sum_{i}^{N_f} \| \sum_{j=1}^{k}(\Delta D_j - \Delta {\hat{D_j}}) - k \times (\Delta D_i - \Delta {\hat{D_i}})\|_1 ,
\end{aligned}
\end{equation}
where $(\Delta D_i - \Delta {\hat{D_i}})$ is the distance of  the pixel-displacement between the predicted flow and the ground-truth flow, and $\sum_{j=1}^{k}$ can be calculated by using convolution which has square kernels with weight of 1. 

Different form general multi-task learning which be applied for assisting with one of the tasks by learning task relationships, our two tasks will be merged for dewarping. We utilize the cosine distance measures the loss in orientation between ground-truth and combined output value of two branch network.
\begin{equation}
L_{cos}= 1 - \cos \theta = 1-\frac{1}{N} \sum_{i}^{N}\frac{\Delta D_i \cdot \Delta \hat{D_i}}{\| \Delta D_i \| \|\Delta \hat{D_i}\|}
\end{equation}

These losses are defined as a linear combination:
\begin{equation}
L=L_B + \alpha L_D + \beta L_{LSC} + \gamma L_{cos} ,
\end{equation}
where $\alpha$, $\beta$ and $\gamma$ are weights associated to $L_D$, $L_{LSC}$ and $L_{cos}$, respectively.

\subsection{Training Details}
In our work, the resolution of input data is 1024 x 960. We train our model on the synthetic dataset of 80,000 images, and none of the documents used in the challenging benchmark dataset proposed by Ma et al.~\cite{ref_9} are used to create the synthetic data for training. The network is trained with Adam optimizer~\cite{other_5}. We set the batch size of 6 and learning rate of $2\times10^{-4}$ which reduced by a factor of 0.5 after each 10 epochs. Our method can produce satisfactory rectified document image in about 30 epochs. We set the hyperparameters as  $\alpha=0.1$, $\beta=0.01$ and $\gamma=0.05$.

\section{Experiments}

\subsection{Datasets}
\begin{figure}
\includegraphics[width=\textwidth]{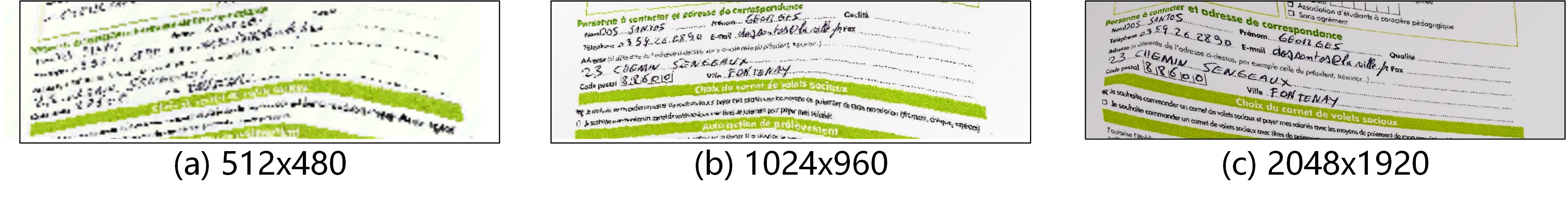}
\caption{Resolution of the synthetic images. The higher the resolution, the more information is retained.  With minor modifications to the model, any resolution can be applied to the training.} \label{fig7}
\end{figure}

Our networks are trained in a supervised manner by synthesizing the distorted document image and the rectified ground-truth. Recently, ~\cite{ref_9,ref_10} generate training dataset in the 3D space to obtain more natural distortion or rich annotations, and use rendering engine to get the finer, realistic details of distorted document image. Although these methods are beneficial for generating more realistic training dataset to improve generalization in real-world images, none of synthetic algorithm could simulate the changeable real-world scenarios. On the other hand, our approach is content-independent which simplifies the difficulty of the dewarping problem and has better performance on rough or unreadable training dataset (as shown in Fig.~\ref{fig7}). Experiments proved that our method still maintains the ability of learning and generalization on real-world images, although the generated training dataset is quite different from the real-world dataset.

For faster and easier synthesizing training dataset, we directly generate distorted document image in 2D mesh.  We warp the scanned document such as receipts, papers and books etc., and then using two functions proposed by~\cite{ref_8} to change the distortion type, such as folds and curves. Meanwhile, we augment the synthetic images by adding various background textures and jitter in the HSV color space. We synthesized 80K images which have the same height and width (i.e., 1024 x 960). Moreover, our ground-truth flow has three channels. For the first two channels, we define the displacement (${\Delta}$x,  ${\Delta}$y) at pixel-level which indicate how far each pixel have to move to reach its position in the undistorted image as the rectified Ground-truth. For the last channel, we represent the foreground or background by using the categories (1 or 0) at pixel-level.

\subsection{Experimental Setup and Results}

We train our network on a synthetic dataset and test on the Ma et al.~\cite{ref_8} benchmark dataset which has various real-world distorted document images. We run our network and post-processing on a NVIDIA TITAN X GPU which processes 10 input images per batch and Intel(R) Xeon(R) CPU E5-2650 v4 which rectifies distorted image by using forward mapping in multiprocessing, respectively. Our implementation takes around 0.67 to 0.72 seconds to process a 1024x960 image. 

\begin{table}
\caption{Comparison of different methods on the Ma et al.~\cite{ref_8} benchmark dataset which has various real-world distorted document images. DocUnet was proposed by Ma et al.~\cite{ref_8}. DewarpNet was proposed by Das and Ma et al.\cite{ref_9} recently, and DewarpNet(ref) is DewarpNet combined with the refinement network to adjust for illumination effects.}\label{tab1}
\centering
\begin{tabular}{M{3cm}|M{3cm}M{3cm}}
\hline\noalign{\smallskip}
{\bfseries Method} &  {\bfseries MS-SSIM} & {\bfseries LD}\\
\noalign{\smallskip}
\hline
\noalign{\smallskip}
DocUnet~\cite{ref_8} &  0.41 & 14.08 \\
DewarpNet\cite{ref_9} &  0.4692 & 8.98 \\
DewarpNet(ref)\cite{ref_9} & {\bfseries 0.4735} & 8.95\\
Our & 0.4361 & {\bfseries 8.50}\\
\hline
\end{tabular}
\end{table}

\begin{figure}
\includegraphics[width=\textwidth]{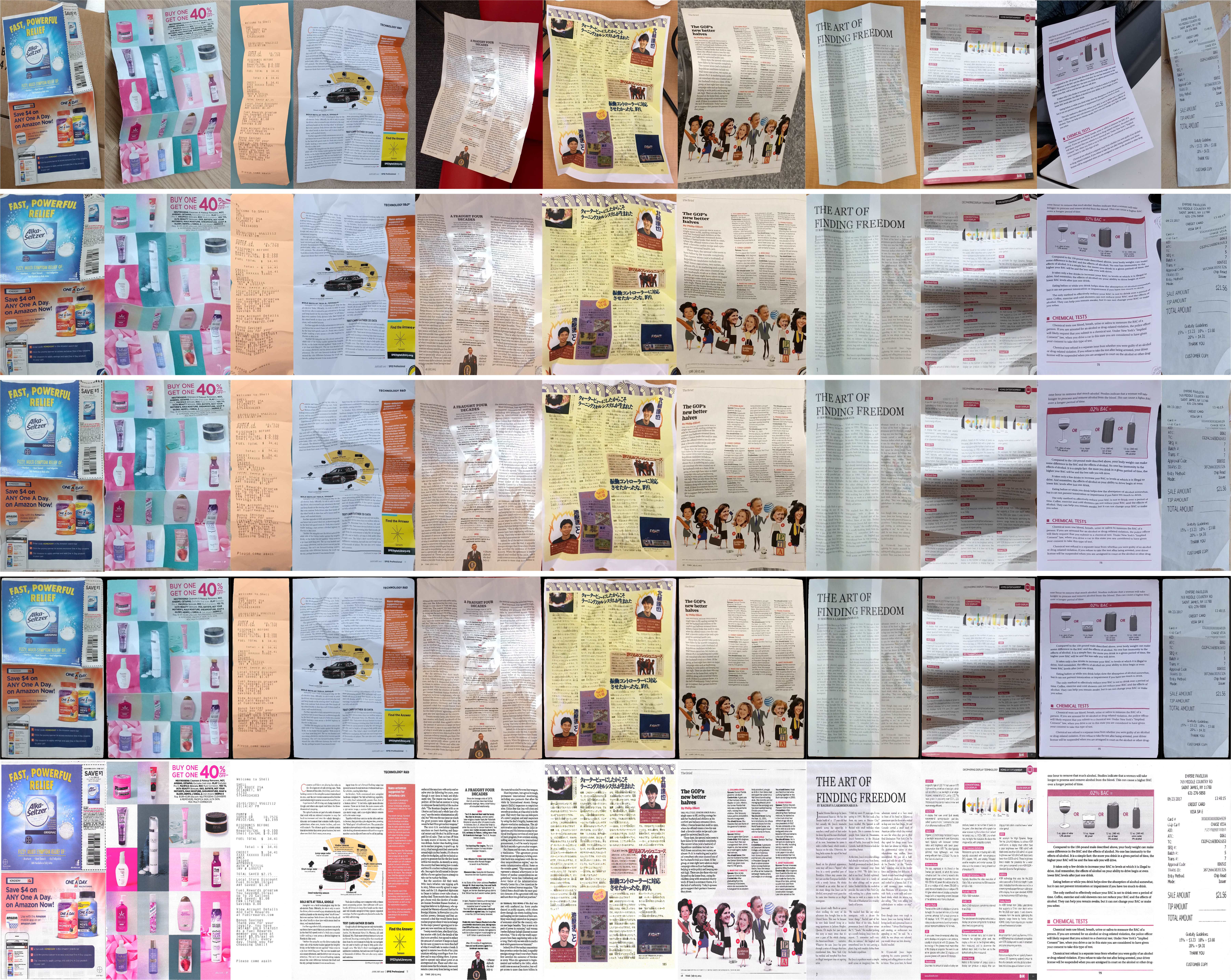}
\caption{Results on the Ma et al.~\cite{ref_8} benchmark dataset. Row 1 : Original distorted images, Row 2 : Results of Ma et al.~\cite{ref_8}, Row 3 : Results of Das and Ma et al.\cite{ref_9},  Row 4 : Results of our method, Row 5 : Scanned images.} \label{fig2}
\end{figure}

\begin{figure}
\includegraphics[width=\textwidth]{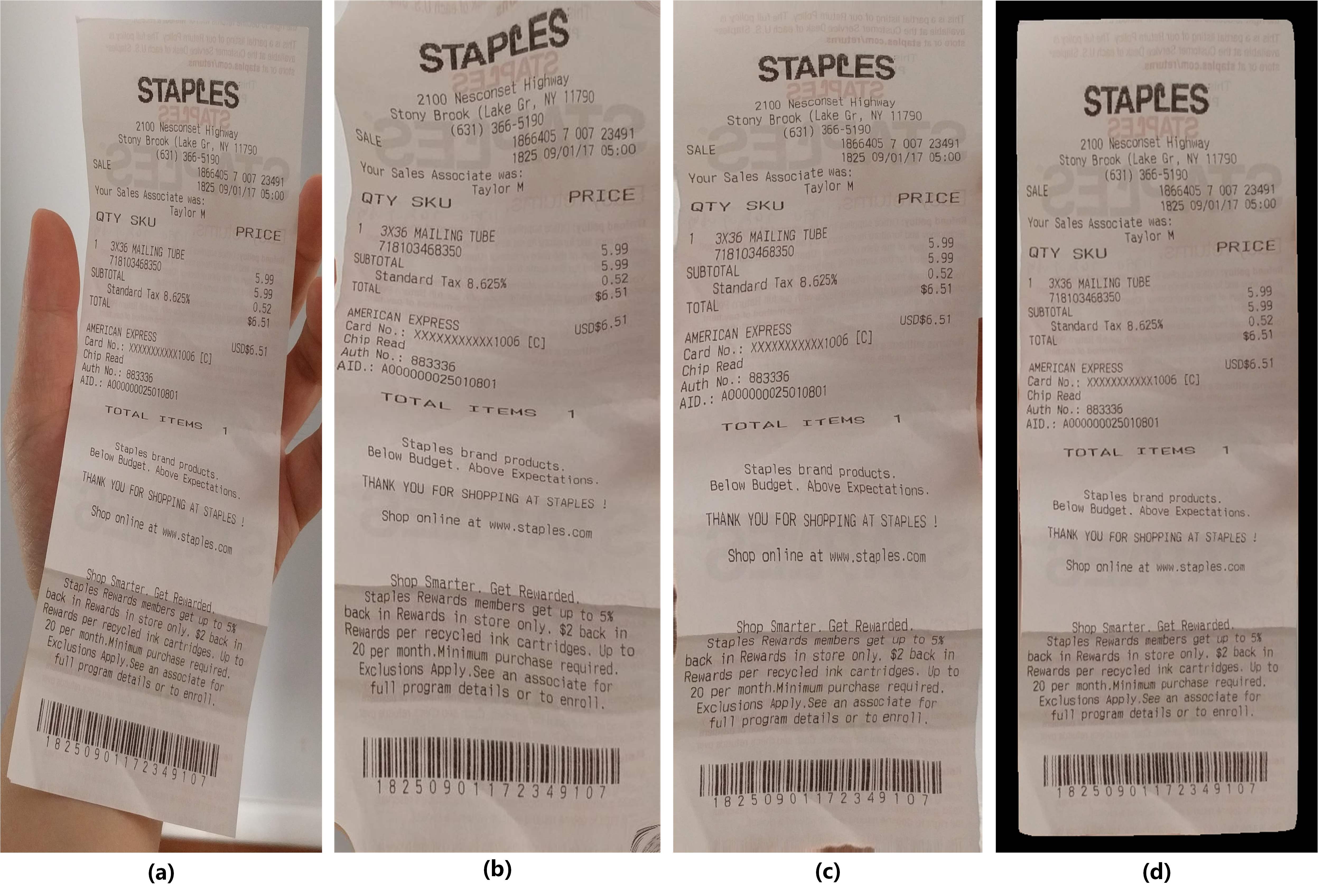}
\caption{Results in details. (a) Original distorted images. (b) Ma et al.~\cite{ref_8}. (c) Das and Ma et al.\cite{ref_9}. (d) Our.} \label{fig8}
\end{figure}

We compare our results with Ma et al.~\cite{ref_8} and Das and Ma et al.\cite{ref_9} on the real-world document images.  Compared with previous method, our proposal can rectify various distortions while removing background and replace it to transparent (the visual comparison is shown in Fig.~\ref{fig2}). As shown in Fig.~\ref{fig8}, our method addresses the difficulties in both rectifying distorted document image and removing background finely. For visually view the process, we don't crop the redundant boundary and retain the original corrected state (No post-cropping, the black edge is the background). 

We use two quantitative evaluation criteria as~\cite{ref_8} which provided the code with default parameters. One of them is Multi-Scale Structural Similarity (MS-SSIM)~\cite{other_6} which evaluate the global similarity between the rectified document images and scanned images in multi-scale. The other one is Local Distortion (LD)~\cite{other_8} which evaluate the local details by computing a dense SIFT flow~\cite{other_7}. The quantitative comparisons between MS-SSIM and LD are shown in Table.~\ref{tab1}. Because our approach is more concerned with how to expand the distorted image than whether the structure is similar, we demonstrate state-of-the-art performance in the quantitative metric of local details.

As shown in  Fig.~\ref{fig5}, Local Smooth Constraint (LSC) can keep the vary continuously from one point to another in a local. With modifying the hyperparameter of $L_D$ and $L_{LSC}$, the effect is similar between the loss functions that measure the element-wise mean squared error and the mean element-wise absolute value difference. In our implementation, the loss function of cosine distance is functionally similar to the pixel-displacement distance, however it can significantly improve the convergence speed. Results in Table.~\ref{tab2} show ablation experiments when we change the hyperparameters of the pixel-displacement and cosine distance. We find that our approach achieves the state-of-the-art performance in terms of local details and overall effect when we set $\alpha=0.1,  \gamma=0.05$, although it's not best in the global similarity (MS-SSIM). Our network is more concerned with expanding the distortions in local and global components, which is no excessive pursuit of the global similarity between the rectified document images and scanned images. Therefore, we can yield the smoother rectified document image and achieves the better performance in overall effect on the various real-world distorted document images.

\begin{table}
\caption{Effect of hyperparameters about the pixel-displacement and cosine distance. $\alpha$ and $\gamma$ represent the hyperparameter of $L_D$ and $L_{cos}$, respectively.}\label{tab2}
\centering
\begin{tabular}{b{4cm}|M{3cm}M{3cm}}
\hline\noalign{\smallskip}
{\bfseries Loss Function} &  {\bfseries MS-SSIM} & {\bfseries LD}\\
\noalign{\smallskip}
\hline
\noalign{\smallskip}
$\alpha=0.1,  \gamma=0.01$ & {\bfseries0.4434} & 8.72 \\
$\alpha=0.1,  \gamma=0.05$ & 0.4361 & {\bfseries 8.50}\\
$\alpha=0.1,  \gamma=0.1$ & 0.4422 & 8.76\\
$\alpha=0.01,  \gamma=0.05$ & 0.4389 & 8.70\\
$\alpha=1,  \gamma=0.05$ & 0.4319 & 9.14\\
\hline
\end{tabular}
\end{table}


\section{Conclusion}
In this paper, we presented a novel framework for rectifying distorted document images using a fully convolutional network for pixel-wise displacement flow estimation and foreground/background classification, so as to address the difficulties in both rectifying distortions and removing background finely. We define a Local Smooth Constraint (LSC) based on the continuum assumption to make the local pixels and global structure of the rectified image to be compact and smooth. Our approach shows the ability of learning and generalization from imperfect virtual training dataset, even if the synthesized dataset is quite different from the real-world dataset. Although our approach has better tradeoff between computational complexity and rectification effect, the edge of partially rectified image was still not neat enough and the speed of calculation still need to be further improved. In the future, we plan to enhance the performance and get rid of the post-processing steps.

\section*{Acknowledgements}
This work has been supported by National Natural Science Foundation of China (NSFC) Grants 61733007, 61573355 and 61721004.

%
%
%
%
\bibliographystyle{splncs04}
\bibliography{reference}

\begin{thebibliography}{10}
\providecommand{\url}[1]{\texttt{#1}}
\providecommand{\urlprefix}{URL }
\providecommand{\doi}[1]{https://doi.org/#1}

\bibitem{other_1}
Amidror, I.: Scattered data interpolation methods for electronic imaging
  systems: a survey. Journal of Electronic Imaging  \textbf{11}(ARTICLE),
  157--76 (2002)

\bibitem{ref_14}
Brown, M.S., Tsoi, Y.C.: Geometric and shading correction for images of printed
  materials using boundary. IEEE Transactions on Image Processing
  \textbf{15}(6),  1544--1554 (2006)

\bibitem{ref_13}
Cao, H., Ding, X., Liu, C.: A cylindrical surface model to rectify the bound
  document image. In: Proceedings Ninth IEEE International Conference on
  Computer Vision. pp. 228--233. IEEE (2003)

\bibitem{other_4}
Chen, L.C., Papandreou, G., Schroff, F., Adam, H.: Rethinking atrous
  convolution for semantic image segmentation. arXiv preprint arXiv:1706.05587
  (2017)

\bibitem{ref_4}
Courteille, F., Crouzil, A., Durou, J.D., Gurdjos, P.: Shape from shading for
  the digitization of curved documents. Machine Vision and Applications
  \textbf{18}(5),  301--316 (2007)

\bibitem{ref_9}
Das, S., Ma, K., Shu, Z., Samaras, D., Shilkrot, R.: Dewarpnet: Single-image
  document unwarping with stacked 3d and 2d regression networks. In:
  Proceedings of the IEEE International Conference on Computer Vision. pp.
  131--140 (2019)

\bibitem{ref_15}
Das, S., Mishra, G., Sudharshana, A., Shilkrot, R.: The common fold: Utilizing
  the four-fold to dewarp printed documents from a single image. In:
  Proceedings of the 2017 ACM Symposium on Document Engineering. pp. 125--128.
  ACM (2017)

\bibitem{ref_7}
Fu, B., Wu, M., Li, R., Li, W., Xu, Z., Yang, C.: A model-based book dewarping
  method using text line detection. In: Proc. 2nd Int. Workshop on Camera Based
  Document Analysis and Recognition, Curitiba, Barazil. pp. 63--70 (2007)

\bibitem{other_2}
He, K., Zhang, X., Ren, S., Sun, J.: Deep residual learning for image
  recognition. In: Proceedings of the IEEE Conference on Computer Vision and
  Pattern Recognition. pp. 770--778 (2016)

\bibitem{other_5}
Kingma, D.P., Ba, J.: Adam: A method for stochastic optimization. arXiv
  preprint arXiv:1412.6980  (2014)

\bibitem{ref_10}
Li, X., Zhang, B., Liao, J., Sander, P.V.: Document rectification and
  illumination correction using a patch-based cnn. ACM Transactions on Graphics
   \textbf{38}(6),  1--11 (2019)

\bibitem{ref_6}
Liang, J., DeMenthon, D., Doermann, D.: Geometric rectification of
  camera-captured document images. IEEE Transactions on Pattern Analysis and
  Machine Intelligence  \textbf{30}(4),  591--605 (2008)

\bibitem{other_7}
Liu, C., Yuen, J., Torralba, A.: Sift flow: Dense correspondence across scenes
  and its applications. IEEE Transactions on Pattern Analysis and Machine
  Intelligence  \textbf{33}(5),  978--994 (2010)

\bibitem{ref_12}
Liu, C., Zhang, Y., Wang, B., Ding, X.: Restoring camera-captured distorted
  document images. International Journal on Document Analysis and Recognition
  \textbf{18}(2),  111--124 (2015)

\bibitem{ref_8}
Ma, K., Shu, Z., Bai, X., Wang, J., Samaras, D.: Docunet: document image
  unwarping via a stacked u-net. In: Proceedings of the IEEE Conference on
  Computer Vision and Pattern Recognition. pp. 4700--4709 (2018)

\bibitem{ref_2}
Meng, G., Wang, Y., Qu, S., Xiang, S., Pan, C.: Active flattening of curved
  document images via two structured beams. In: Proceedings of the IEEE
  Conference on Computer Vision and Pattern Recognition. pp. 3890--3897 (2014)

\bibitem{ref_17}
Ramanna, V., Bukhari, S.S., Dengel, A.: Document image dewarping using deep
  learning. In: International Conference on Pattern Recognition Applications
  and Methods (2019)

\bibitem{ref_5}
Tian, Y., Narasimhan, S.G.: Rectification and 3d reconstruction of curved
  document images. In: Proceedings of the IEEE Conference on Computer Vision
  and Pattern Recognition. pp. 377--384. IEEE (2011)

\bibitem{ref_11}
Tsoi, Y.C., Brown, M.S.: Multi-view document rectification using boundary. In:
  Proceedings of the IEEE Conference on Computer Vision and Pattern
  Recognition. pp.~1--8. IEEE (2007)

\bibitem{ref_3}
Wada, T., Ukida, H., Matsuyama, T.: Shape from shading with interreflections
  under a proximal light source: Distortion-free copying of an unfolded book.
  International Journal of Computer Vision  \textbf{24}(2),  125--135 (1997)

\bibitem{other_3}
Wang, P., Chen, P., Yuan, Y., Liu, D., Huang, Z., Hou, X., Cottrell, G.:
  Understanding convolution for semantic segmentation. In: IEEE winter
  conference on applications of computer vision. pp. 1451--1460. IEEE (2018)

\bibitem{other_6}
Wang, Z., Simoncelli, E.P., Bovik, A.C.: Multiscale structural similarity for
  image quality assessment. In: The Thrity-Seventh Asilomar Conference on
  Signals, Systems \& Computers. vol.~2, pp. 1398--1402. Ieee (2003)

\bibitem{ref_16}
Xing, Y., Li, R., Cheng, L., Wu, Z.: Research on curved chinese document
  correction based on deep neural network. In: International Symposium on
  Computational Intelligence and Design. vol.~2, pp. 342--345. IEEE (2018)

\bibitem{other_8}
You, S., Matsushita, Y., Sinha, S., Bou, Y., Ikeuchi, K.: Multiview
  rectification of folded documents. IEEE Transactions on Pattern Analysis and
  Machine Intelligence  \textbf{40}(2),  505--511 (2017)

\bibitem{ref_1}
Zhang, L., Zhang, Y., Tan, C.: An improved physically-based method for
  geometric restoration of distorted document images. IEEE Transactions on
  Pattern Analysis and Machine Intelligence  \textbf{30}(4),  728--734 (2008)

\end{thebibliography}

\end{document}